\documentclass[runningheads]{llncs}
\usepackage[T1]{fontenc}
\usepackage{graphicx,verbatim}
\usepackage{amsmath}
\usepackage{amsmath,amssymb}
\usepackage{multirow} 
\usepackage{booktabs}
\usepackage{array}
\usepackage{xcolor}
\usepackage{subfigure}
\usepackage{caption}
\usepackage[table]{xcolor}
\usepackage{colortbl}
\usepackage{capt-of}   
\usepackage{hyperref}
\usepackage{bbding}

\begin{document}
\title{IDRL: An Individual-Aware Multimodal Depression-Related Representation Learning Framework for Depression Diagnosis}
\titlerunning{IDRL for Multimodal Depression Diagnosis}
%

\author{Chongxiao Wang\inst{1,2}$^\dagger$ 
\and
Junjie Liang\inst{1,2}$^\dagger$ 
\and
Peng Cao\inst{1,2,3}$^{\Envelope}$ 
\and
Jinzhu Yang\inst{1,2,3} 
\and
Osmar R. Zaiane\inst{4}}
\footnotetext[1]{$\dagger$ Chongxiao Wang and Junjie Liang contributed equally to this work.}

\authorrunning{C. Wang et al.}
%
\institute{School of Computer Science and Engineering, Northeastern University, Shenyang, China \and
Key Laboratory of Intelligent Computing in Medical Image of Ministry of Education, Northeastern University, Shenyang, China \and
National Frontiers Science Center for Industrial Intelligence and Systems Optimization, Shenyang, China \\
\email{caopengneu@gmail.com} \and
Alberta Machine Intelligence Institute, University of Alberta, Edmonton, Canada}
  
\maketitle              
\begin{abstract}

Depression is a severe mental disorder, and reliable identification plays a critical role in early intervention and treatment.
Multimodal depression detection aims to improve diagnostic performance by jointly modeling complementary information from multiple modalities.
Recently, numerous multimodal learning approaches have been proposed for depression analysis; however, these methods suffer from the following limitations: 1) inter-modal inconsistency and depression-unrelated interference, where depression-related cues may conflict across modalities while substantial irrelevant content obscures critical depressive signals, and 2) diverse individual depressive presentations, leading to individual differences in modality and cue importance that hinder reliable fusion.
To address these issues, we propose Individual-aware Multimodal Depression-related Representation Learning Framework (IDRL) for robust depression diagnosis.
Specifically, IDRL 1) disentangles multimodal representations into a modality-common depression space, a modality-specific depression space, and a depression-unrelated space to enhance modality alignment while suppressing irrelevant information, and 2) introduces an individual-aware modality-fusion module (IAF) that dynamically adjusts the weights of disentangled depression-related features based on their predictive significance, thereby achieving adaptive cross-modal fusion for different individuals.
Extensive experiments demonstrate that IDRL achieves superior and robust performance for multimodal depression detection. The code is available at \url{https://anonymous.4open.science/r/IDRL-code-6153/}.

\keywords{Collaborative decoupling  \and Individual-Awareness \and Multimodal depression \and Modality-common and Modality-specific features.}

\end{abstract}
\section{Introduction}
\begin{figure}[!t]
\centering
\includegraphics[width=0.7\columnwidth]{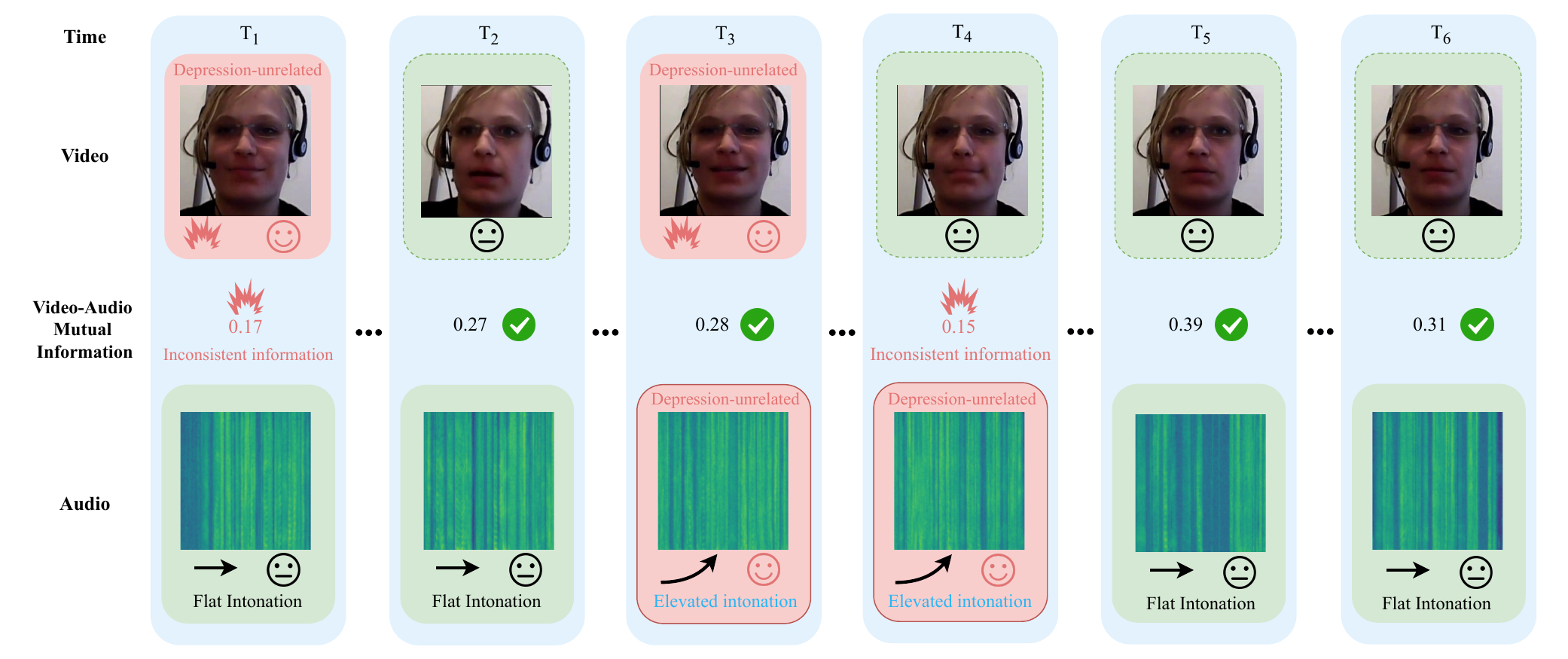}
\caption{
The video-audio mutual information score of each time segment is calculated on the AVEC-2014 depression dataset. Higher mutual information scores indicate stronger consistency across modalities within the corresponding segments.
}
\label{fig1}
\end{figure}

Depression is a severe global mental disorder, yet diagnosis still depends heavily on time-consuming and subjective clinical judgment, motivating automated prediction to support early intervention and decision-making.

Early studies mainly focused on unimodal cues (e.g., speech, facial behaviors, or text), yet a single modality may fail to capture comprehensive depressive signals~\cite{huang2024depression,nassibi2022depression,wang2024facialpulse}.
With the complementarity of heterogeneous signals, multimodal methods have been explored to integrate cross-modal information for more robust depression diagnosis~\cite{zhang2024multimodal,cheng2022multimodal}. 
Despite the progress, two challenges still hinder reliable multimodal depression detection.

\textbf{(1) Inter-modal inconsistency and depression-unrelated interference.}
Modalities can be consistent overall yet conflict locally, and multimodal recordings inevitably contain substantial depression-unrelated segments.
As shown in Fig.~\ref{fig1}, audio and video exhibit high mutual information (MI) in most segments~\cite{de2006scoring}, but pronounced conflicts at T1/T4 and depression-unrelated behaviors (e.g., natural facial expressions, elevated vocal tone, and faster speech at T1/T3/T4) may obscure critical depressive signals, motivating segment-level inconsistency-aware and interference-robust fusion.
\textbf{(2) Individual differences in depressive presentations.}
Depressive cues vary substantially across individuals, such that the diagnostic salience of different modalities and behavioral cues differs from one subject to another~\cite{fried2015depression}.

To tackle these issues, we propose an \emph{Individual-aware Multimodal Depression-related Representation Learning} framework (IDRL) for robust depression diagnosis (Fig.~\ref{fig2}). 
IDRL consists of a depression-representation disentanglement module (DRD) and an individual-aware modality-fusion module (IAF). 
DRD constructs a modality-common depression space, a modality-specific depression space, and a depression-unrelated space via modality-wise collaborative disentanglement, thereby improving modality-consistent alignment while suppressing depression-unrelated interference. 
To model individual differences during fusion, the individual-aware modality-fusion module dynamically adjusts the weights of depression-related features in both the modality-common and modality-specific depression spaces according to their predictive significance, enabling adaptive cross-modal fusion for each individual.

We evaluate IDRL on two benchmark datasets, and experiments validate its effectiveness for multimodal depression detection.
Our contributions are three-fold:
\begin{itemize}
    \item \textbf{Multimodal depression-related representation learning via disentanglement.} 
    We propose DRD to explicitly model modality-common and modality-specific depression-related information while separating depression-unrelated features.
    \item \textbf{Individual-aware depression modeling.} 
    We design the IAF to adaptively reweight depression-related features in the modality-common and specific depression spaces.
    \item \textbf{Extensive evaluations on multiple datasets.} 
    Experiments and analyses verify the effectiveness and robustness of IDRL across different modality combinations.
\end{itemize}

\section{Method}

\subsection{Overview}
\begin{figure*}[!t]
\centering
\includegraphics[width=0.8\textwidth]{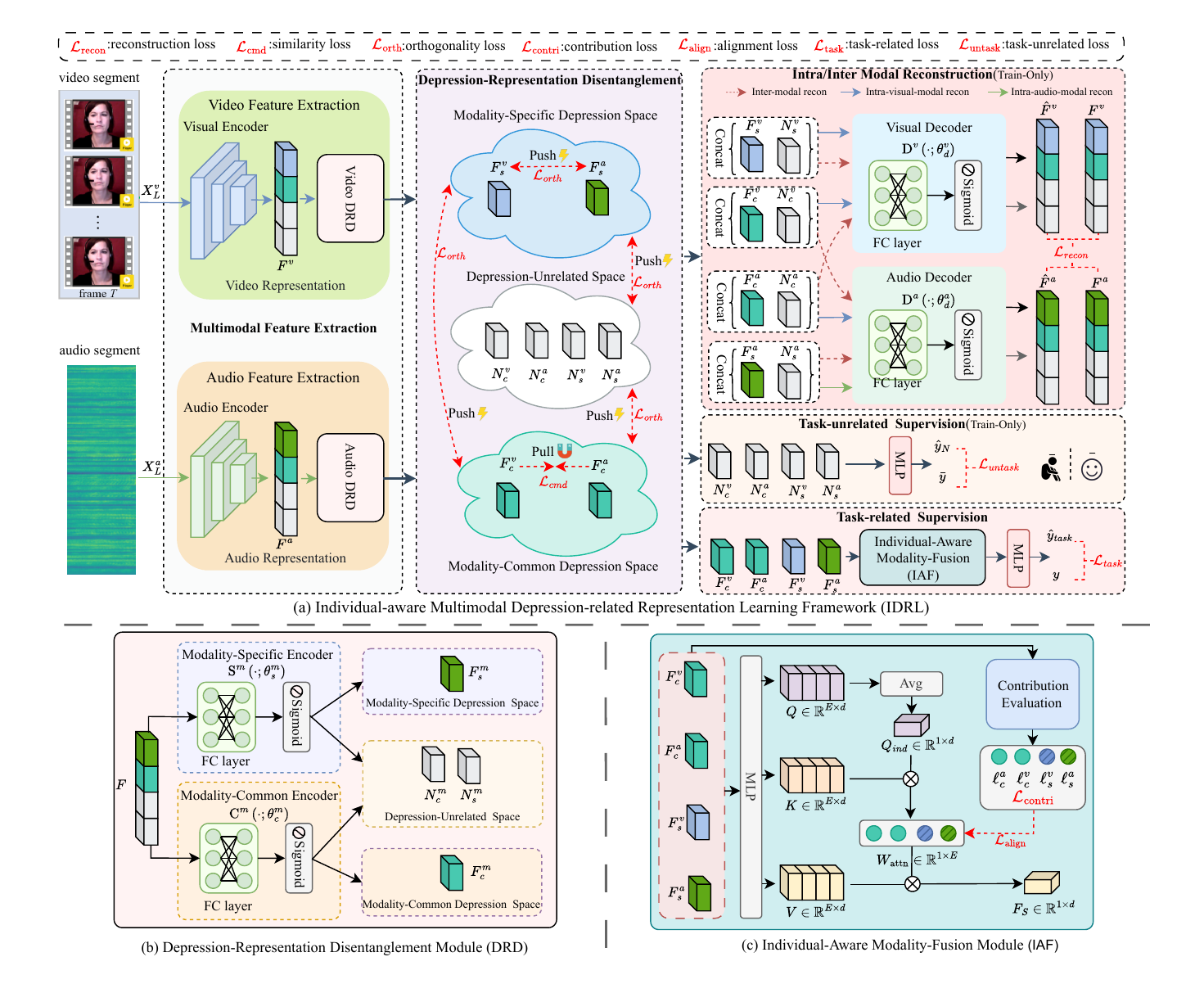}
\caption{(a)The proposed Individual-aware Multimodal Depression-related Representation Learning Framework (IDRL) consists of three key stages: multimodal feature extraction, depression-representation disentanglement, and feature fusion and prediction.(b)Depression Representation Disentanglement Module (DRD).(c)Individual-Aware Modality-Fusion Module (IAF).
}
\label{fig2}
\end{figure*}
We propose an Individual-aware Multimodal Depression-related Representation Learning framework (IDRL) for depression diagnosis. Our framework is modality-agnostic. In this work, we evaluate two instantiations: (video, audio) on AVEC and (text, image) on Twitter.
As illustrated in Fig.~\ref{fig2}(a), we first extract aligned visual and acoustic features over $L$ temporal segments, denoted as $F^{v}$ and $F^{a}$, respectively.
IDRL then consists of two key modules: a Depression-Representation Disentanglement (DRD) module and an individual-aware modality-fusion module (IAF) for depression prediction.




\subsection{Depression-Representation Disentanglement}

\subsubsection{Depression-Representation Disentanglement Module (DRD)}
As illustrated in Fig.~\ref{fig2}(b), DRD disentangles multimodal representations into a modality-common depression space, a modality-specific depression space, and a depression-unrelated space.
For each modality $m\in\{v,a\}$ with extracted feature $F^{m}$, we employ two encoders: a modality-common depression encoder $\mathrm{E}^{m}_\mathrm{com}$ and a modality-specific depression encoder $\mathrm{E}^{m}_\mathrm{spe}$, i.e.,  
$[F_{c}^{m},\, N_{c}^{m}]=\mathrm{E}^{m}_\mathrm{com}(F^{m};\theta_{c}^{m})$ and $[F_{s}^{m},\, N_{s}^{m}]=\mathrm{E}^{m}_\mathrm{spe}(F^{m};\theta_{s}^{m})$,
where $F_{c}^{m}$ and $F_{s}^{m}$ denote modality-common and modality-specific depression-related features, while $N_{c}^{m}$ and $N_{s}^{m}$ denote the corresponding depression-unrelated features.

\subsubsection{Guidance for Disentanglement}

\paragraph{Reconstruction loss.}

We reconstruct $F^{m}$ from the disentangled components using a decoder $\mathrm{D}^{m}$. 
Specifically, we perform self- and cross-modal reconstruction by swapping the modality-common depression-related feature, i.e., 
$\hat{F}_{\text{self}}^{m}=\mathrm{D}^{m}\!\left(N_c^{m}\oplus N_s^{m}\oplus F_s^{m}\oplus F_c^{m};\theta_{d}^{m}\right)$ and 
$\hat{F}_{\text{cross}}^{m}=\mathrm{D}^{m}\!\left(N_c^{m}\oplus N_s^{m}\oplus F_s^{m}\oplus F_c^{\bar{m}};\theta_{d}^{m}\right)$, 
where $\bar{m}$ denotes the other modality and $\oplus$ is concatenation. 
Reconstruction preserves the information of $F^{m}$ in the disentangled features, and cross-modal reconstruction further encourages cross-modal interaction via the modality-common depression space.
We use the mean squared error:
\begin{equation}
\small
\mathcal{L}_{\text{recon}}=\frac{1}{|C|}\sum_{m\in\{v,a\}}
\left(
\left\|F^{m}-\hat{F}_{\text{self}}^{m}\right\|_{2}^{2}+
\left\|F^{m}-\hat{F}_{\text{cross}}^{m}\right\|_{2}^{2}
\right),
\end{equation}
where $|C|$ is the number of modalities.

\paragraph{Similarity loss.}
To align modality-common depression-related features across modalities, we use Central Moment Discrepancy (CMD)~\cite{zellinger2017central}:
\begin{equation}
\small
\mathcal{L}_{\text{cmd}}=\operatorname{CMD}_{K}(F_{c}^{v},F_{c}^{a})=
\frac{1}{|b-a|}\left\|\mathbb{E}(F_{c}^{v})-\mathbb{E}(F_{c}^{a})\right\|_{2}
+\sum_{k=2}^{K}\frac{1}{|b-a|^{k}}
\left\|C_{k}(F_{c}^{v})-C_{k}(F_{c}^{a})\right\|_{2},
\end{equation}
where $C_{k}(\cdot)$ denotes the $k$-th central moment, $K$ is the maximum order,
and $|b-a|$ is the feature range in the mini-batch ($a=\min(\cdot)$, $b=\max(\cdot)$).

\paragraph{Orthogonal loss.}
To encourage separation among modality-common depression-related, modality-specific depression-related, and depression-unrelated components, we apply soft orthogonal regularization~\cite{wen2013feasible}:
\begin{equation}
\small
\mathcal{L}_{\text{orth}}=
\sum_{m\in\{v,a\}}\left(
\left\|\left(F_{c}^{m}\right)^{\top}F_{s}^{m}\right\|_{F}^{2}
+\left\|\left(F_{c}^{m}\right)^{\top}N_{c}^{m}\right\|_{F}^{2}
+\left\|\left(F_{s}^{m}\right)^{\top}N_{s}^{m}\right\|_{F}^{2}
\right).
\end{equation}

\paragraph{Diagnosis-relevant Loss.}

We predict the depression score from the fused depression-related feature $F_S$ as
$\hat{y}_{\mathrm{task}}=\psi_{\mathrm{MLP}}^{S}(F_S;\theta^{S})$,
and minimize
$\mathcal{L}_{\mathrm{task}}=\frac{1}{R}\sum_{i=1}^{R}(y_i^{\mathrm{reg}}-\hat{y}_{\mathrm{task}})^2$,
where $y^{\mathrm{reg}}$ denotes the ground-truth depression score.

To suppress depression information in the unrelated space, we stack
$N=[(N_c^v)^\top,(N_c^a)^\top,(N_s^v)^\top,(N_s^a)^\top]\in\mathbb{R}^{E\times d}$,
and obtain $\hat{y}^N=\psi_{\mathrm{MLP}}^{N}(N;\theta^{N})$.
For AVEC-2014, we use the auxiliary binary label $y^{\mathrm{aux}}$ (derived from BDI-II$\ge$14) and minimize the BCE with the reversed label
$\bar{y}^{\mathrm{aux}}=1-y^{\mathrm{aux}}$:
$\mathcal{L}_{\mathrm{untask}}=\mathrm{BCE}(\hat{y}^N,\bar{y}^{\mathrm{aux}})$,
which makes $N$ anti-predictive of depression and thus prevents depression-related cues from leaking into $N$.

\subsection{Individual-Aware Modality Fusion}

\subsubsection{Individual-Aware Modality Fusion Module}

As shown in Fig.~\ref{fig2}(c), we fuse disentangled depression-related features via an individual-aware weighting mechanism. 
We stack modality-common and modality-specific depression-related features from both modalities to form 
$S=\big[\left(F_{c}^{v}\right)^{\top},\left(F_{c}^{a}\right)^{\top},\left(F_{s}^{v}\right)^{\top},\left(F_{s}^{a}\right)^{\top}\big]\in\mathbb{R}^{E\times d}$, 
where $E$ is the number of stacked depression-related features. 
We then compute $Q=SW_{D}^{Q}$, $K=SW_{D}^{K}$, and $V=SW_{D}^{V}$, with $Q,K,V\in\mathbb{R}^{E\times d}$. 
To obtain an individual query (a sample-specific individual-level summary), we average $Q$ along the stacked dimension as $Q_{\mathrm{ind}}=\frac{1}{E}\sum_{i=1}^{E}Q_{i}\in\mathbb{R}^{1\times d}$. 
The attention weights over disentangled features are $W_{\mathrm{attn}}=\mathrm{softmax}\!\left(\frac{Q_{\mathrm{ind}}K^{\top}}{\sqrt{d}}\right)\in\mathbb{R}^{1\times E}$, 
and the fused depression-related representation is $F_{S}=W_{\mathrm{attn}}V\in\mathbb{R}^{1\times d}$.

\subsubsection{Guidance for Individual-Aware Fusion}
\paragraph{Contribution loss.}

We evaluate the predictive contribution of each disentangled depression-related feature using an auxiliary MLP-based classifier (a single fully connected layer followed by a Sigmoid) as the contribution evaluation head:
$\hat{y}^{m}_{u}=\mathrm{MLP}_{\mathrm{aux}}(F^{m}_{u})$, where $u\in\{c,s\}$.
We then compute the binary cross-entropy loss
$\ell_{u}^{m}=-\big[y^{\mathrm{aux}}\log\hat{y}_{u}^{m}+(1-y^{\mathrm{aux}})\log(1-\hat{y}_{u}^{m})\big]$.
For AVEC-2014 (regression), we derive the auxiliary binary label $y^{\mathrm{aux}}\in\{0,1\}$ from the BDI-II score
($y^{\mathrm{aux}}=1$ if $\mathrm{BDI\text{-}II}\ge 14$, indicating at least mild depressive symptoms) solely for contribution estimation~\cite{wang2013psychometric}, without altering the regression objective.
The overall contribution loss is obtained by summing losses over modalities and spaces:
$\mathcal{L}_{\text{contri}}=\sum_{m\in\{v,a\}}\sum_{u\in\{c,s\}}\ell_{u}^{m}$.

\paragraph{Alignment loss.}
To encourage attention weights to be consistent with predictive contribution, we align $W_{\mathrm{attn}}$ with the ranking induced by $\ell$ using a pairwise margin ranking loss~\cite{burges2005learning}:
\begin{equation}
\small
\mathcal{L}_{\mathrm{align}}
=\frac{1}{3|C|}
\sum_{i=1}^{|C|}
\sum_{j=1}^{|U|}
\sum_{\substack{p=1 \\ p \neq i}}^{|C|}
\sum_{\substack{q=1 \\ q \neq j}}^{|U|}
\max\!\left(
0,\,
\mathbb{1}\!\left[\ell^{m_i}_{u_j} < \ell^{m_p}_{u_q}\right]
\left(
W^{m_p}_{u_q,\mathrm{attn}}-W^{m_i}_{u_j,\mathrm{attn}}+\epsilon
\right)
\right),
\end{equation}
where $\mathbb{1}[\cdot]$ is the indicator function, which equals 1 if the condition is satisfied and -1 otherwise, and we set $\epsilon=0.05$.

\paragraph{Overall Loss.}
The model is optimized by minimizing
\begin{equation}
\small
\mathcal{L}_{\text{total}}
=
\underbrace{(\mathcal{L}_{\text{task}}+\mathcal{L}_{\text{untask}})}_{\text{diagnosis-relevant loss}}
+\alpha \underbrace{(\mathcal{L}_{\text{orth}}+\mathcal{L}_{\text{cmd}}+\mathcal{L}_{\text{recon}})}_{\text{disentanglement loss}}
+\beta \underbrace{(\mathcal{L}_{\text{align}}+\mathcal{L}_{\text{contri}})}_{\text{individual-aware loss}} ,
\end{equation}
where we set $\alpha=0.7$ and $\beta=0.5$.

\section{Experiments} 
\subsection{Datasets and Evaluation Metrics.}
\textit{Datasets:} We conduct experiments on two public benchmarks. AVEC-2014~\cite{valstar2014avec} provides interview-based audio--video recordings with BDI-II scores as regression targets. Twitter~\cite{gui2019cooperative} contains 1,402 depressed users and 1,402 control users; since only a small portion of tweets include images, we apply an attention mask in the Transformer to handle missing image regions.
\textit{Evaluation metrics:} For AVEC-2014, we report MAE and RMSE for BDI score prediction. For Twitter, we report Accuracy and Macro-F1.

\subsection{Implementation Details.}
All experiments are conducted in PyTorch on a single NVIDIA RTX A30 GPU.
For AVEC-2014, we use LI-FPN~\cite{pan2024spatial} and a CNN encoder with MFCC inputs as the video and audio backbones, respectively. For Twitter, we adopt RoBERTa and CLIP (ViT) as the text and image backbones, respectively. 
For a fair comparison, we follow the backbone settings used in prior work whenever available; otherwise, we re-implement baselines under the same backbone and training protocol as ours.
Standard pre-processing is applied (AVEC-2014: 15 fps and aligned $224{\times}224$ face crops; Twitter: a 32-step sliding window with padding).
IDRL is optimized with Adam using a learning rate of $1{\times}10^{-3}$ and a batch size of 16; early stopping is applied, and the checkpoint with the best validation performance is reported.

\setlength{\extrarowheight}{1.5pt}
\begin{table}[t]
\centering
\small

\begin{minipage}[t]{0.49\columnwidth}
\centering
\caption{Comparison of Our Method and State-Of-The-Art Works on the Test Set of AVEC-2014 Dataset}
\label{table1}
\resizebox{\linewidth}{!}{
\begin{tabular}{ccccc}
\toprule
Modalities & Type & Methods                & MAE↓   & RMSE↓  \\
\hline
\multirow{5}{*}{A} & \multirow{4}{*}{ND}
                     & MRELBP-DCNN\cite{he2018automated}
                     & 8.19      & 9.99      \\
                  &  & FTA-Net\cite{li2023fta}
                     & 7.31      & 9.60      \\ 
                 &  & WavDepressionNet\cite{niu2023wavdepressionnet}
                     & 6.60      & 8.61      \\ \cline{2-5}
                  & \multirow{1}{*}{D}  & IDRL(Audio)
                     & 6.48      & 8.28      \\
\hline
\multirow{5}{*}{V} & \multirow{4}{*}{ND}
                  & SubAttn-V\cite{wei2022multi}
                     & 7.84      & 10.70     \\
                  &  & MTDAN\cite{zhang2023mtdan}
                     & 6.36      & 7.94      \\
                  &  & MDN\cite{de2021mdn}
                     & 6.06      & 7.66      \\ \cline{2-5}
                  & \multirow{1}{*}{D} & IDRL(Video)
                     & 5.94      & 7.57      \\
\hline
\multirow{7}{*}{A+V} & \multirow{3}{*}{ND}
                     & MEN-ADF\cite{zhang2024novel}
                     & 7.02      & 9.39      \\
                  &  & GMM-ELM\cite{williamson2014vocal}
                     & 6.31      & 8.13      \\
                  &  & FDHH-AV\cite{jan2017artificial}
                     & 6.14      & 7.45      \\
\cline{2-5}
                  & \multirow{4}{*}{D}
                     & MISA\cite{hazarika2020misa}
                     & 7.12      & 8.95     \\
                  &  & Disentangled-MER\cite{yang2022disentangled}
                     & 6.68      & 8.41      \\
                  &  & TDRL\cite{zhou2025triple}
                     & 5.97      & 7.63      \\
                  &  & IDRL(Audio+Video)
                     & 5.83      & 7.34      \\
\bottomrule
\end{tabular}
}\vspace{1mm}

\footnotesize{A \textit{and} V \textit{are audio and video modalities. }
↓ \textit{indicates that smaller is better.}}
\end{minipage}
\hfill
\begin{minipage}[t]{0.49\columnwidth}
\centering
\caption{Comparison of Our Method and State-Of-The-Art Works on the Twitter Dataset using 5-Fold Cross-
Validation}
\label{table2}
\resizebox{\linewidth}{!}{
\begin{tabular}{ccccc}
\toprule
Modalities & Type & Methods        & Accuracy↑      & F1↑  \\
\hline
\multirow{3}{*}{T} & \multirow{2}{*}{ND}
                     & T-LSTM\cite{baytas2017patient}
                     & 0.854      & 0.849\\
                  &  & DDSM-RL\cite{gui2019depression}
                     & 0.870      & 0.872 \\
                  &  & EmoBerta\cite{kim2021emoberta}
                     & 0.861      & 0.864 \\ \cline{2-5}
                  & \multirow{1}{*}{D}  & IDRL (Text)
                     & 0.916      & 0.903      \\
\hline
\multirow{10}{*}{T+V} & \multirow{6}{*}{ND}
                     & MTAL\cite{an2020multimodal}
                     & 0.842      & 0.842 \\
                  &  & MTAN\cite{cheema2021role}
                     & 0.869      & 0.871 \\
                  &  & CMAM\cite{gui2019cooperative}
                     & 0.900      & 0.900 \\
                  &  & TEMT\cite{bucur2023s}
                     & 0.896      & 0.898 \\
                  &  & MTAAN\cite{cheng2022multimodal}
                     & 0.921      & 0.887 \\ \cline{2-5}
                  & \multirow{4}{*}{D}
                     & MISA\cite{hazarika2020misa}
                     & 0.864     & 0.866 \\
                  &  &Disentangled-MER\cite{yang2022disentangled}
                     & 0.883      & 0.887 \\
                  &  & TDRL\cite{zhou2025triple}
                     & 0.913      & 0.904 \\
                  &  & IDRL (Text+Image)
                     & 0.943      & 0.932 \\
\bottomrule
\end{tabular}
}\vspace{1mm}

\footnotesize{T \textit{and} V \textit{are text and image modalities. }
ND \textit{refers to Non-Decoupled. }
D \textit{refers to Decoupled. }
↑ \textit{denotes higher is better.}}
\end{minipage}

\end{table}

\subsubsection{Comparable Methods and Comparative Results.}
We compare IDRL with representative state-of-the-art methods, including unimodal and multimodal baselines as well as disentanglement-based approaches (Tables~\ref{table1}--\ref{table2}). Overall, the quantitative results show that IDRL consistently outperforms existing methods across datasets and modality combinations, demonstrating its effectiveness under different modality settings (video--audio on AVEC-2014 and text--image on Twitter). Specifically, IDRL achieves the best performance on AVEC-2014 and improves over the strongest decoupled baseline, and also attains the best results on Twitter. These gains are attributed to disentangling modality-common and modality-specific depression-related representations while suppressing depression-unrelated information, together with individual-aware reweighting for adaptive multimodal fusion.

\subsection{Ablation Study}

\subsubsection{Analysis of the proposed components.}

We conduct ablation experiments on AVEC-2014 to evaluate the contribution of DRD and IAF. “Concat$\rightarrow$MLP/Trans.” concatenates multimodal features and uses an MLP/Transformer predictor, while “MLP/Trans. fuse” applies an MLP/Transformer fusion block over the DRD outputs $(F_{c}^{m}, F_{s}^{m})$ before regression. 
As shown in Table~\ref{tab:ablation_components}, performance improves consistently as each component is incorporated into the baseline, and the full IDRL achieves the best results. This suggests that DRD improves depression-related consistency by suppressing depression-unrelated interference, and IAF further strengthens fusion through adaptive reweighting.

\subsubsection{Effects of Loss Functions}

\begin{figure}[!t]
\centering
\subfigure[t-SNE visualization of disentangled depression-related and unrelated features]{
    \includegraphics[width=0.4\columnwidth]{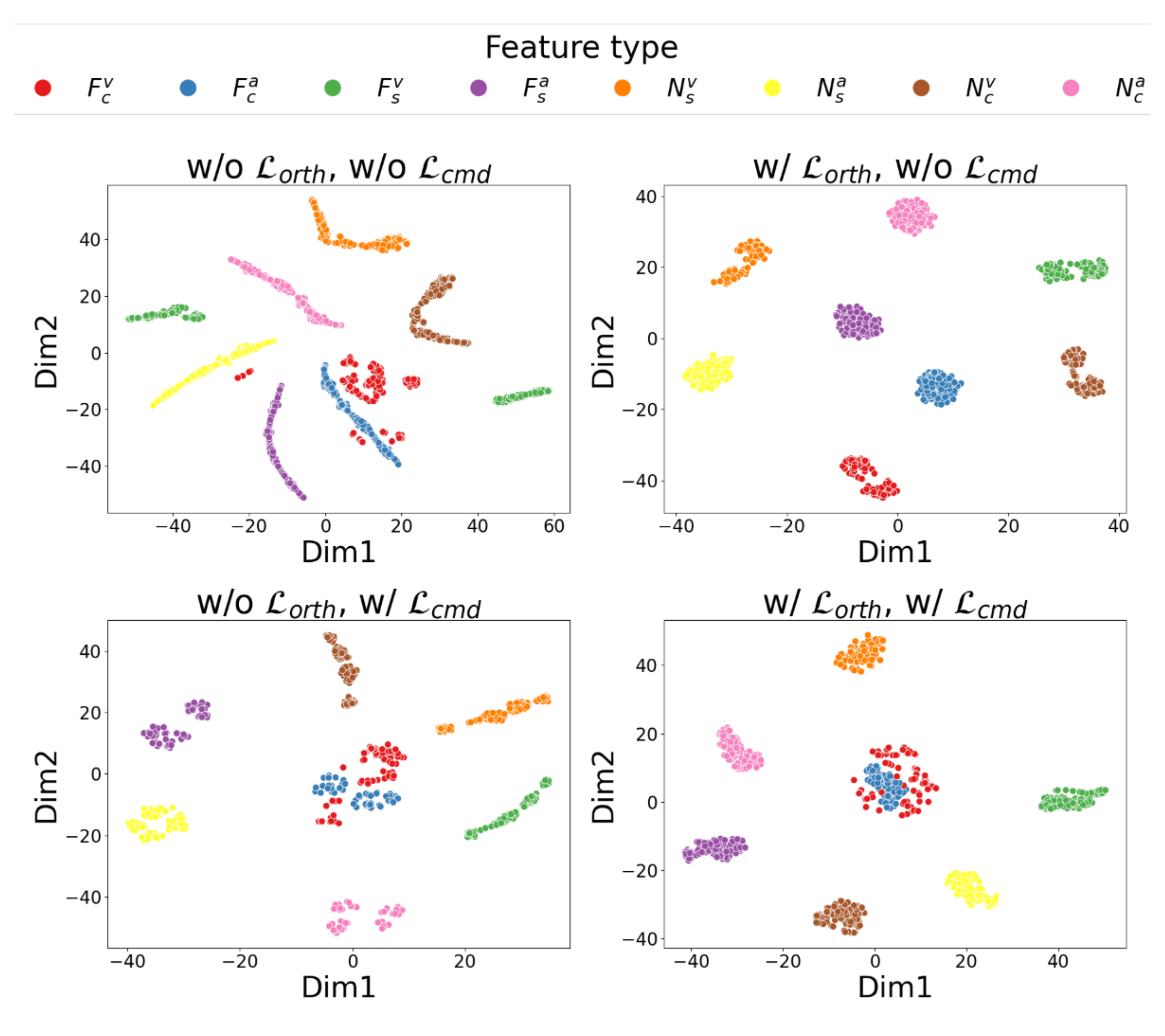}
}
\hfill
\subfigure[Grad-CAM++ visualization of the impact of task-unrelated loss on depression prediction]{
    \includegraphics[width=0.4\columnwidth]{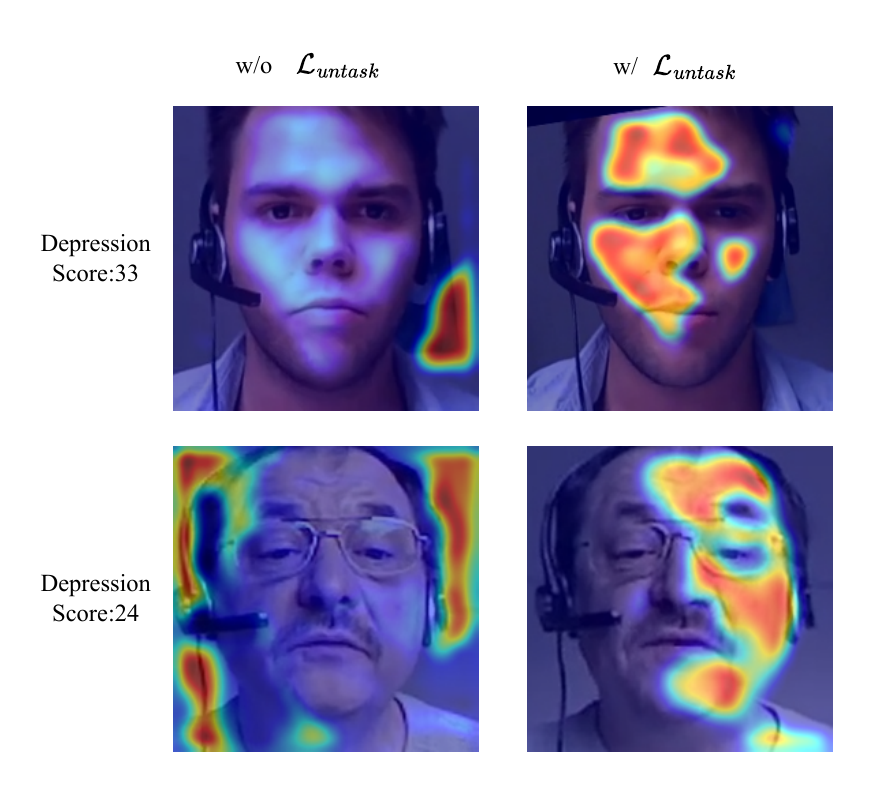}
}
\caption{Visualization of (a) disentangled depression-related and unrelated features and (b) the effect of task-unrelated loss on depression prediction.}
\label{fig6}
\end{figure}

As reported in Table~\ref{table7}, removing $\mathcal{L}{\text{orth}}$ or $\mathcal{L}{\text{cmd}}$ yields the largest performance drops, underscoring the importance of enforcing subspace separation and cross-modal consistency in DRD. The remaining terms ($\mathcal{L}{\text{untask}}$, $\mathcal{L}{\text{align}}$, and $\mathcal{L}{\text{contri}}$) also provide consistent gains by reducing depression-irrelevant leakage and stabilizing sample-adaptive fusion. Fig.~\ref{fig6} offers qualitative evidence: without $\mathcal{L}{\text{orth}}/\mathcal{L}{\text{cmd}}$, modality-common, modality-specific, and depression-unrelated embeddings overlap more in t-SNE, while introducing $\mathcal{L}{\text{untask}}$ shifts Grad-CAM++ activations from background/peripheral regions to facial areas, indicating more effective suppression of depression-unrelated cues.

\begin{table*}[t]
\centering
\scriptsize
\setlength{\tabcolsep}{2.4pt}
\renewcommand{\arraystretch}{0.90}

\begin{minipage}[t]{0.48\textwidth}
\centering
\caption{Ablation of proposed components on AVEC-2014.}
\label{tab:ablation_components}
\resizebox{\linewidth}{!}{%
\begin{tabular}{p{0.72\linewidth}c}
\toprule
Setting & MAE$\downarrow$/RMSE$\downarrow$ \\
\midrule
Baseline (Concat$\!\rightarrow\!$MLP) & 8.46/10.66 \\
Baseline (Concat$\!\rightarrow\!$Trans.) & 7.52/9.41 \\
\midrule
Baseline+DRD (MLP fuse) & 7.19/9.27 \\
Baseline+DRD (Trans. fuse) & 6.33/8.44 \\
\midrule
\textbf{IDRL (Baseline+DRD+IAF)} & \textbf{5.83/7.34} \\
\bottomrule
\end{tabular}%
}
\vspace{0.35mm}
\end{minipage}%
\hfill%
\begin{minipage}[t]{0.50\textwidth}
\centering
\caption{Ablation of the loss function on AVEC-2014.}
\label{table7}
\resizebox{\linewidth}{!}{%
\begin{tabular}{ccccccc|cc}
\toprule
$\mathcal{L}_{\text{orth}}$ & $\mathcal{L}_{\text{cmd}}$ & $\mathcal{L}_{\text{untask}}$ &
$\mathcal{L}_{\text{align}}$ & $\mathcal{L}_{\text{contri}}$ & $\mathcal{L}_{\text{recon}}$ &
$\mathcal{L}_{\text{task}}$ & MAE$\downarrow$ & RMSE$\downarrow$ \\
\midrule
$\times$ & \checkmark & \checkmark & \checkmark & \checkmark & \checkmark & \checkmark & 7.28 & 9.04 \\
\checkmark & $\times$ & \checkmark & \checkmark & \checkmark & \checkmark & \checkmark & 7.01 & 8.66 \\
\checkmark & \checkmark & $\times$ & \checkmark & \checkmark & \checkmark & \checkmark & 6.94 & 8.47 \\
\checkmark & \checkmark & \checkmark & $\times$ & \checkmark & \checkmark & \checkmark & 6.21 & 7.69 \\
\checkmark & \checkmark & \checkmark & \checkmark & $\times$ & \checkmark & \checkmark & 6.13 & 7.51 \\
\checkmark & \checkmark & \checkmark & \checkmark & \checkmark & $\times$ & \checkmark & 6.08 & 7.42 \\
\checkmark & \checkmark & \checkmark & \checkmark & \checkmark & \checkmark & \checkmark & \textbf{5.83} & \textbf{7.34} \\
\bottomrule
\end{tabular}%
}
\vspace{0.35mm}
\scriptsize{The best result is highlighted in bold.}
\end{minipage}
\end{table*}




\section{Conclusion}
We propose an individual-aware multimodal depression-related representation learning framework (IDRL), for robust depression diagnosis. IDRL disentangles multimodal features into modality-common, modality-specific, and depression-unrelated spaces to alleviate inter-modal inconsistency and suppress irrelevant interference, and further performs individual-aware fusion by adaptively weighting informative cues to accommodate individual differences. Extensive experiments on AVEC-2014 and Twitter show that IDRL consistently outperforms representative state-of-the-art methods, with ablation and visualization analyses validating the contribution of each component.

\bibliographystyle{elsarticle-num}
\bibliography{ref}   

@inproceedings{burges2005learning,
  title={Learning to rank using gradient descent},
  author={Burges, Chris and Shaked, Tal and Renshaw, Erin and Lazier, Ari and Deeds, Matt and Hamilton, Nicole and Hullender, Greg},
  booktitle={Proceedings of the 22nd international conference on Machine learning},
  pages={89--96},
  year={2005}
}

@article{pan2024spatial,
  title={Spatial--temporal attention network for depression recognition from facial videos},
  author={Pan, Yuchen and Shang, Yuanyuan and Liu, Tie and Shao, Zhuhong and Guo, Guodong and Ding, Hui and Hu, Qiang},
  journal={Expert systems with applications},
  volume={237},
  pages={121410},
  year={2024},
  publisher={Elsevier}
}

@article{jan2017artificial,
  title={Artificial intelligent system for automatic depression level analysis through visual and vocal expressions},
  author={Jan, Asim and Meng, Hongying and Gaus, Yona Falinie Binti A and Zhang, Fan},
  journal={IEEE Transactions on Cognitive and Developmental Systems},
  volume={10},
  number={3},
  pages={668--680},
  year={2017},
  publisher={IEEE}
}

@inproceedings{baytas2017patient,
  title={Patient subtyping via time-aware LSTM networks},
  author={Baytas, Inci M and Xiao, Cao and Zhang, Xi and Wang, Fei and Jain, Anil K and Zhou, Jiayu},
  booktitle={Proceedings of the 23rd ACM SIGKDD international conference on knowledge discovery and data mining},
  pages={65--74},
  year={2017}
}

@article{cheema2021role,
  title={On the role of images for analyzing claims in social media},
  author={Cheema, Gullal S and Hakimov, Sherzod and M{\"u}ller-Budack, Eric and Ewerth, Ralph},
  journal={arXiv preprint arXiv:2103.09602},
  year={2021}
}

@article{kim2021emoberta,
  title={Emoberta: Speaker-aware emotion recognition in conversation with roberta},
  author={Kim, Taewoon and Vossen, Piek},
  journal={arXiv preprint arXiv:2108.12009},
  year={2021}
}

@inproceedings{gui2019depression,
  title={Depression detection on social media with reinforcement learning},
  author={Gui, Tao and Zhang, Qi and Zhu, Liang and Zhou, Xu and Peng, Minlong and Huang, Xuanjing},
  booktitle={China National Conference on Chinese Computational Linguistics},
  pages={613--624},
  year={2019},
  organization={Springer}
}

@article{he2018automated,
  title={Automated depression analysis using convolutional neural networks from speech},
  author={He, Lang and Cao, Cui},
  journal={Journal of biomedical informatics},
  volume={83},
  pages={103--111},
  year={2018},
  publisher={Elsevier}
}

@inproceedings{wei2022multi,
  title={Multi-modal depression estimation based on sub-attentional fusion},
  author={Wei, Ping-Cheng and Peng, Kunyu and Roitberg, Alina and Yang, Kailun and Zhang, Jiaming and Stiefelhagen, Rainer},
  booktitle={European Conference on Computer Vision},
  pages={623--639},
  year={2022},
  organization={Springer}
}

@article{de2021mdn,
  title={MDN: A deep maximization-differentiation network for spatio-temporal depression detection},
  author={de Melo, Wheidima Carneiro and Granger, Eric and Lopez, Miguel Bordallo},
  journal={IEEE transactions on affective computing},
  volume={14},
  number={1},
  pages={578--590},
  year={2021},
  publisher={IEEE}
}

@article{zellinger2017central,
  title={Central moment discrepancy (CMD) for domain-invariant representation learning},
  author={Zellinger, Werner and Grubinger, Thomas and Lughofer, Edwin and Natschl{\"a}ger, Thomas and Saminger-Platz, Susanne},
  journal={arXiv preprint arXiv:1702.08811},
  year={2017}
}

@article{wen2013feasible,
  title={A feasible method for optimization with orthogonality constraints},
  author={Wen, Zaiwen and Yin, Wotao},
  journal={Mathematical Programming},
  volume={142},
  number={1},
  pages={397--434},
  year={2013},
  publisher={Springer}
}

@article{zhang2023mtdan,
  title={MTDAN: A lightweight multi-scale temporal difference attention networks for automated video depression detection},
  author={Zhang, Shiqing and Zhang, Xingnan and Zhao, Xiaoming and Fang, Jiangxiong and Niu, Mingyue and Zhao, Ziping and Yu, Jun and Tian, Qi},
  journal={IEEE transactions on affective computing},
  volume={15},
  number={3},
  pages={1078--1089},
  year={2023},
  publisher={IEEE}
}

@inproceedings{an2020multimodal,
  title={Multimodal topic-enriched auxiliary learning for depression detection},
  author={An, Minghui and Wang, Jingjing and Li, Shoushan and Zhou, Guodong},
  booktitle={proceedings of the 28th international conference on computational linguistics},
  pages={1078--1089},
  year={2020}
}

@inproceedings{valstar2014avec,
  title={Avec 2014: 3d dimensional affect and depression recognition challenge},
  author={Valstar, Michel and Schuller, Bj{\"o}rn and Smith, Kirsty and Almaev, Timur and Eyben, Florian and Krajewski, Jarek and Cowie, Roddy and Pantic, Maja},
  booktitle={Proceedings of the 4th international workshop on audio/visual emotion challenge},
  pages={3--10},
  year={2014}
}

@inproceedings{yang2022disentangled,
  title={Disentangled representation learning for multimodal emotion recognition},
  author={Yang, Dingkang and Huang, Shuai and Kuang, Haopeng and Du, Yangtao and Zhang, Lihua},
  booktitle={Proceedings of the 30th ACM International Conference on Multimedia},
  pages={1642--1651},
  year={2022}
}

@article{cheng2022multimodal,
  title={Multimodal time-aware attention networks for depression detection},
  author={Cheng, Ju Chun and Chen, Arbee LP},
  journal={Journal of Intelligent Information Systems},
  volume={59},
  number={2},
  pages={319--339},
  year={2022},
  publisher={Springer}
}

@article{zhang2024multimodal,
  title={Multimodal Sensing for Depression Risk Detection: Integrating Audio, Video, and Text Data},
  author={Zhang, Zhenwei and Zhang, Shengming and Ni, Dong and Wei, Zhaoguo and Yang, Kongjun and Jin, Shan and Huang, Gan and Liang, Zhen and Zhang, Li and Li, Linling and others},
  journal={Sensors},
  volume={24},
  number={12},
  pages={3714},
  year={2024},
  publisher={MDPI}
}

@article{huang2024depression,
  title={Depression recognition using voice-based pre-training model},
  author={Huang, Xiangsheng and Wang, Fang and Gao, Yuan and Liao, Yilong and Zhang, Wenjing and Zhang, Li and Xu, Zhenrong},
  journal={Scientific Reports},
  volume={14},
  number={1},
  pages={12734},
  year={2024},
  publisher={Nature Publishing Group UK London}
}

@inproceedings{wang2024facialpulse,
  title={FacialPulse: An Efficient RNN-based Depression Detection via Temporal Facial Landmarks},
  author={Wang, Ruiqi and Huang, Jinyang and Zhang, Jie and Liu, Xin and Zhang, Xiang and Liu, Zhi and Zhao, Peng and Chen, Sigui and Sun, Xiao},
  booktitle={Proceedings of the 32nd ACM International Conference on Multimedia},
  pages={311--320},
  year={2024}
}

@inproceedings{hazarika2020misa,
  title={Misa: Modality-invariant and-specific representations for multimodal sentiment analysis},
  author={Hazarika, Devamanyu and Zimmermann, Roger and Poria, Soujanya},
  booktitle={Proceedings of the 28th ACM international conference on multimedia},
  pages={1122--1131},
  year={2020}
}

@article{wang2013psychometric,
  title={Psychometric properties of the Beck Depression Inventory-II: a comprehensive review},
  author={Wang, Yuan-Pang and Gorenstein, Clarice},
  journal={Revista brasileira de psiquiatria},
  volume={35},
  number={4},
  pages={416--431},
  year={2013},
  publisher={SciELO Brasil}
}

@article{zhou2025triple,
  title={Triple disentangled representation learning for multimodal affective analysis},
  author={Zhou, Ying and Liang, Xuefeng and Chen, Han and Zhao, Yin and Chen, Xin and Yu, Lida},
  journal={Information Fusion},
  volume={114},
  pages={102663},
  year={2025},
  publisher={Elsevier}
}

@inproceedings{williamson2014vocal,
  title={Vocal and facial biomarkers of depression based on motor incoordination and timing},
  author={Williamson, James R and Quatieri, Thomas F and Helfer, Brian S and Ciccarelli, Gregory and Mehta, Daryush D},
  booktitle={Proceedings of the 4th international workshop on audio/visual emotion challenge},
  pages={65--72},
  year={2014}
}

@article{zhang2024novel,
  title={A novel multimodal depression diagnosis approach utilizing a new hybrid fusion method},
  author={Zhang, Xiufeng and Li, Bingyi and Qi, Guobin},
  journal={Biomedical Signal Processing and Control},
  volume={96},
  pages={106552},
  year={2024},
  publisher={Elsevier}
}

@inproceedings{gui2019cooperative,
  title={Cooperative multimodal approach to depression detection in twitter},
  author={Gui, Tao and Zhu, Liang and Zhang, Qi and Peng, Minlong and Zhou, Xu and Ding, Keyu and Chen, Zhigang},
  booktitle={Proceedings of the AAAI conference on artificial intelligence},
  volume={33},
  number={01},
  pages={110--117},
  year={2019}
}

@inproceedings{bucur2023s,
  title={It’s just a matter of time: Detecting depression with time-enriched multimodal transformers},
  author={Bucur, Ana-Maria and Cosma, Adrian and Rosso, Paolo and Dinu, Liviu P},
  booktitle={European Conference on Information Retrieval},
  pages={200--215},
  year={2023},
  organization={Springer}
}

@article{de2006scoring,
  title={A scoring function for learning Bayesian networks based on mutual information and conditional independence tests.},
  author={De Campos, Luis M and Friedman, Nir},
  journal={Journal of Machine Learning Research},
  volume={7},
  number={10},
  year={2006}
}

@article{nassibi2022depression,
  title={Depression diagnosis using machine intelligence based on spatiospectrotemporal analysis of multi-channel EEG},
  author={Nassibi, Amir and Papavassiliou, Christos and Atashzar, S Farokh},
  journal={Medical \& Biological Engineering \& Computing},
  volume={60},
  number={11},
  pages={3187--3202},
  year={2022},
  publisher={Springer}
}

@article{fried2015depression,
  title={Depression sum-scores don’t add up: why analyzing specific depression symptoms is essential},
  author={Fried, Eiko I and Nesse, Randolph M},
  journal={BMC medicine},
  volume={13},
  number={1},
  pages={72},
  year={2015},
  publisher={Springer}
}

@inproceedings{li2023fta,
  title={Fta-net: A frequency and time attention network for speech depression detection},
  author={Li, Qifei and Wang, Dong and Ren, Yiming and Gao, Yingming and Li, Ya},
  booktitle={Proc. INTERSPEECH},
  volume={2023},
  pages={1723--1727},
  year={2023}
}

@article{niu2023wavdepressionnet,
  title={Wavdepressionnet: Automatic depression level prediction via raw speech signals},
  author={Niu, Mingyue and Tao, Jianhua and Li, Yongwei and Qin, Yong and Li, Ya},
  journal={IEEE Transactions on Affective Computing},
  volume={15},
  number={1},
  pages={285--296},
  year={2023},
  publisher={IEEE}
}

\end{document}